\documentclass{article}
\usepackage{spconf,amsmath,graphicx}
\usepackage[titlenumbered,ruled]{algorithm2e}
\usepackage{xspace}
\usepackage{balance}
\usepackage{color}
\usepackage{pifont}
\newcommand{\cmark}{\ding{51}}%
\newcommand{\xmark}{\ding{55}}%

\usepackage{enumitem}
\setlist{nosep, leftmargin=14pt}

\usepackage{mwe} 

\title{Objective-Dependent Uncertainty Driven Retinal Vessel Segmentation}
\name{Suraj Mishra, Danny Z. Chen, X. Sharon Hu}
\address{Department of Computer Science and Engineering, University of Notre Dame, Indiana, USA}

\begin{document}
%
\maketitle

\begin{abstract}
From diagnosing neovascular diseases to detecting white matter lesions, accurate tiny vessel segmentation in fundus images is critical. Promising results for accurate vessel segmentation have been known. However, their effectiveness in segmenting tiny vessels is still limited. In this paper, we study retinal vessel segmentation by incorporating tiny vessel segmentation into our framework for the overall accurate vessel segmentation. To achieve this, we propose a new deep convolutional neural network (CNN) which divides vessel segmentation into two separate objectives. Specifically, we consider the overall accurate vessel segmentation and tiny vessel segmentation as two individual objectives. Then, by exploiting the objective-dependent (homoscedastic) uncertainty, we enable the network to learn both objectives simultaneously. Further, to improve the individual objectives, we propose: (a) a vessel weight map based auxiliary loss for enhancing tiny vessel connectivity (i.e., improving tiny vessel segmentation), and (b) an enhanced encoder-decoder architecture for improved localization (i.e., for accurate vessel segmentation). Using 3 public retinal vessel segmentation datasets (CHASE\_DB1, DRIVE, and STARE), we verify the superiority of our proposed framework in segmenting tiny vessels (8.3\% average improvement in sensitivity) while achieving better area under the receiver operating characteristic curve (AUC) compared to state-of-the-art methods.  
\end{abstract}


\section{Introduction}
\label{sec:intro}
Retinal vasculature analysis in fundus images plays a vital role in early diagnosis of various diseases, e.g., glaucoma, age related macular degeneration, arteriosclerosis, and multiple sclerosis~\cite{chasedb1,mou,li,drive,orlando}. Clinically, for neovascular disease diagnosis and cerebral small vessel disease studies, accurate tiny vessel segmentation is critical~\cite{mou}. Specifically, the low arteriolar-to-venular ratio (AVR) is attributed to the presence of white matter lesions along with predictive features of several cardiovascular diseases and diabetic retinopathy~\cite{mou,avr}.

Since manual segmentation of retinal vessels is a labor- and time-intensive task, significant research has been conducted for automatic segmentation of retinal vessels. In recent years, convoutional neural network (CNN) based approaches for retinal vessel segmentation~\cite{mou,xia,yishuo} have outperformed traditional approaches using hand-craft features~\cite{chasedb1}. In one such CNN based early work~\cite{yishuo}, vessel segmentation was viewed as a multi-class segmentation problem by introducing additional labels. A multi-scale network-following-network model was utilized in~\cite{xia,wu_new} for enhanced feature extraction. Even with some success, these methods failed to segment tiny vessels well. Recently, Mou~\textit{et al.}~\cite{mou} proposed a densely dilated network along with a probability regularized random walk based post-processing for tiny vessel segmentation along with broken vessel rectification. However, the performance of such a post-processing was limited by the network's output and preconceived vessel shape approximation. 

\begin{figure}[tb]
    \vspace*{-0.2in}
    \includegraphics[width=0.384\textwidth]{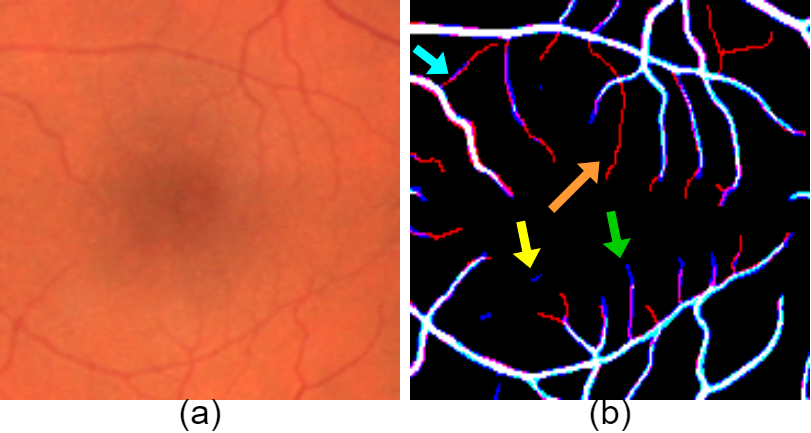}
    \centering
    \vspace*{-0.125in}
    \caption{(a) A sample fundus image region. (b) Ground truth is shown in Red channel. Segmentation outputs without and with the auxiliary loss are shown in Green and Blue channels, respectively. The magenta (red+blue) regions are vessels segmented using the auxiliary loss only. Green arrow highlights tiny vessel segmentation. Yellow arrow highlights false positives. Orange arrow shows unsegmented tiny vessels and Cyan arrow shows broken vessels.}
    \label{fig:task_div}  
    \vspace*{-0.12in}
\end{figure}

Since tiny vessels (small and/or thin vessels) constitute only a small portion of the total image area (e.g., in the training images of the DRIVE dataset~\cite{drive}, the vessel regions constitute less than $10\%$ of the total image area and tiny vessels of 3 pixel width or smaller are less than $50\%$ of the total vessel pixels), the network may \textit{ignore} them during training. In order to better learn tiny vessels, Mishra~\textit{et al.}~\cite{mishra} proposed a data-aware deep supervision technique by analyzing vessel widths and introducing an additional auxiliary loss. However, introduction of such an auxiliary loss made the network highly sensitive, resulting in lower overall accuracy. An example case highlighting segmentation output with and without the auxiliary loss is given in Fig.~\ref{fig:task_div}. As shown in the figure, in the absence of the auxiliary loss, the network tends to ignore tiny vessels, causing it to attain higher accuracy metrics except sensitivity. On the other hand, inclusion of the auxiliary loss increases sensitivity but introduces false positive and broken vessels, causing reduction in the overall accuracy. Such contrasting behaviors of the model can be attributed to the aleatoric uncertainty or the genuine stochasticity \cite{kendall} associated with the retinal vessel data which under the current training protocol may be difficult to capture. 

We hypothesize that an uncertainty driven Bayesian model is more suitable for feature generalization, making the model capable of capturing contrasting but shared representations. When learning a shared representation, uncertainty driven multi-objective learning can improve learning efficiency and prediction accuracy~\cite{cipolla}. By sharing domain information required for different objectives, better feature generalization can be achieved in multi-objective learning. Existing methods~\cite{cipolla,bischke} have successfully explored uncertainty in multi-objective learning for vision tasks such as simultaneous depth prediction and instance segmentation. 

To achieve accurate tiny vessel segmentation using Bayes-ian modeling, we exploit the idea of uncertainty driven multi-objective learning in solving the vessel segmentation problem, by strategically dividing the task of vessel segmentation into two separate objectives. Specifically, we consider accurate vessel segmentation and tiny vessel segmentation as two separate objectives with shared representation. We explore the homoscedastic uncertainty between these two objectives to obtain accurate tiny vessel segmentation. Further, we propose techniques to improve the individual objectives, as follows. (a) To enhance connectivity of tiny vessels, we introduce additional penalty on tiny vessel segmentation. Using a vessel weight map based loss, such a penalty is realized which reduces vessel breakage. (b) To improve vessel localization, we design an enhanced encoder-decoder architecture. By appending coarser feature maps to a targeted upsampling stage of the network, our proposed architecture can generate accurate vessel segmentation. Our framework is verified using 3 public vessel segmentation datasets (CHASE\_DB1, DRIVE, and STARE), showing its efficacy.

\section{Methodology}
\label{sec:method}
Our proposed framework (shown in Fig.~\ref{fig:overall}) has two major components: (1) uncertainty modelling, and (2) individual objective improvement. Section~\ref{ssec:unc_mod} presents our objective uncertainty driven modelling. Section~\ref{ssec:obj-imp} describes the techniques used to improve the individual objectives.

\subsection{Uncertainty Modelling}
\label{ssec:unc_mod}
Bayesian modelling can be divided into two major categories: \textit{epistemic uncertainty} captures uncertainty in the model due to lack of training data; \textit{aleatoric uncertainty} captures information stochasticity in the data~\cite{kendall}. Further, aleatoric uncertainty is subdivided into: \textit{heteroscedastic} or input data dependent; \textit{homoscedastic} or output objective dependent. We first describe our two objectives for accurate tiny vessel segmentation. Then we explore output objective uncertainty and its extension to vessel segmentation.

\noindent\textbf{Two Objectives}: Our framework treats accurate vessel segmentation and tiny vessel segmentation as two separate objectives. For accurate vessel segmentation, the main network output is used to generate the main loss correlating with our first objective. An auxiliary loss is used as the second objective representing tiny vessel segmentation. We use a data-aware approach for auxiliary output location selection as in~\cite{mishra}, where the average vessel width is used to match the layer-wise effective receptive field ($LERF$) of the network to determine the \textit{preeminent layer} for best representing the target vessel features. Essentially, in this work we use uncertainty to \textit{learn} the relative confidence between the main and the auxiliary objectives, to determine a good trade-off, reflecting their associated uncertainty.

\begin{figure}[tb]
    \vspace*{-0.2in}
    \includegraphics[width=0.48\textwidth]{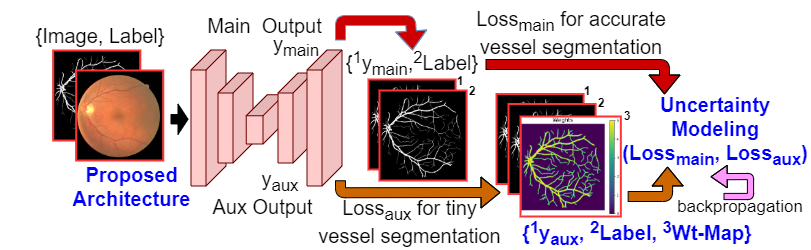}
    \centering
    \vspace*{-0.2in}
    \caption{Our proposed framework. Major contributions of this work are highlighted in blue color.}
    \label{fig:overall} 
    \vspace*{-0.1in}
\end{figure}

\noindent\textbf{Objective Uncertainty}: For uncertainty modeling, taking inspiration from Bayesian neural network \cite{kendall,denker}, we formulate the network input-output relation as a probability distribution, i.e., for a deterministic input, the network's output is probabilistic containing observation noise. Then using the negative log likelihood, we formulate the optimization objective or the loss function to capture the network output uncertainty. Consider a CNN with $\theta$ parameters. The network output for an input $i$ and target $o$ with $c$ classes can be represented as $f^\theta(i)$. As in~\cite{cipolla}, we examine the pixel-wise likelihood by squashing a scaled version of the network output through a softmax function $\mathcal{SM}$ as $p(o|f^\theta(i), \sigma) = \mathcal{SM}(\frac{1}{\sigma^2}f^\theta(i))$, where $\sigma$ is a positive scalar determining how `uniform' (flat) the discrete distribution is. $\sigma$ can be interpreted as uncertainty or entropy associated with the discrete Boltzmann distribution. Now the log likelihood of the output becomes: $\log(p(o=c|f^\theta(i), \sigma)) = \frac{1}{\sigma^2}f^\theta_c(i) - \log \sum_{c'} \exp{\frac{1}{\sigma^2}f^\theta_{c'}(i)}$. By rearrangement: $\frac{1}{\sigma^2} \log \exp(f^\theta_c(i)) - \frac{1}{\sigma^2} \log \sum_{c'} \exp (f^\theta_{c'}(i))$  $- \log\sum_{c'} \exp{\frac{1}{\sigma^2}f^\theta_{c'}(i)} + \frac{1}{\sigma^2} \log \sum_{c'} \exp (f^\theta_{c'}(i))$ becomes $ \frac{1}{\sigma^2} \log \frac{\exp(f^\theta_c(i))}{\sum_{c'}\exp (f_\theta^{c'}(i))} - \log \frac{\sum_{c'} \exp{\frac{1}{\sigma^2}f^\theta_{c'}(i)}}{\big(\sum_{c'} \exp (f^\theta_{c'}(i))\big)^\frac{1}{\sigma^2}}.$ The explicit approximation, $\big(\sum_{c'} \exp (f^\theta_{c'}(i))\big)^\frac{1}{\sigma^2} \approx  \frac{1}{\sigma} \sum_{c'} \exp{\frac{1}{\sigma^2}f^\theta_{c'}(i)}$, results in $\frac{1}{\sigma^2} \log \mathcal{SM}(f^\theta_c(i)) - \log\sigma$. Rewriting $ \mathcal{L}(\theta, \sigma) = \frac{1}{\sigma^2}\mathcal{L}(\theta) + \log\sigma$, where the \textit{negative} log likelihood of the softmax as the loss function to be minimized. Here, $\sigma$ is the observation noise parameter related to network's output uncertainty (as measured in entropy), which is learned along with the network weights ($\theta$) during training. 

\noindent\textbf{Homoscedastic Uncertainty and Loss Functions}: Each of our two objectives, corresponds to a separate loss associated with different network outputs ($f^{\theta_{main}}(i)$ or $y_{main}$ and $f^{\theta_{aux}}(i)$ or $y_{aux}$, as shown in Fig.~\ref{fig:overall}). Using the output objective uncertainty capturing loss function derived in the previous section, the overall loss function combining both losses becomes: $\mathcal{L}(\theta_{main},\theta_{aux}, \sigma_{main}, \sigma_{aux})$ = $-\log(p(o|f^{\theta_{main}}(i), o|f^{\theta_{aux}}(i))) = \frac{1}{\sigma_{main}^2}\mathcal{L}_{main}(\theta_{main}) + \frac{1}{\sigma_{aux}^2}\mathcal{L}_{aux}(\theta_{aux}) + \log\sigma_{main} + \log\sigma_{aux}$. Such a combination of weighted losses forces the network to learn both the objectives simultaneously. $\sigma_{main}$ and $\sigma_{aux}$ can be viewed as two learnable control `knobs', which the network learns to adjust in order to generate a minima satisfying both the objectives. Since $\mathcal{L}_{aux}$ only utilizes vessel weight map (see Section~\ref{ssec:obj-imp}) for loss generation, $\mathcal{L}_{main} \neq \mathcal{L}_{aux}$. 

\begin{figure}[tb]
    \vspace*{-0.085in}
    \includegraphics[width=0.45\textwidth]{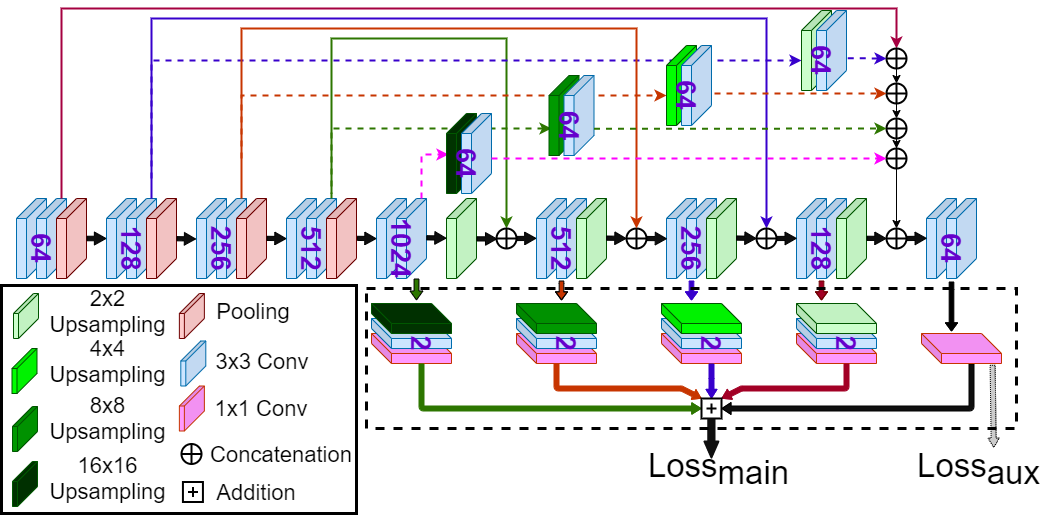}
    \centering
    \vspace*{-0.1in}
    \caption{Our encoder-decoder architecture with ILCs.}
    \label{fig:mod-icc}  
    \vspace*{-0.1in}
\end{figure}

\begin{figure*}[tb]
    \vspace*{-0.2in}
    \includegraphics[width=0.99\textwidth]{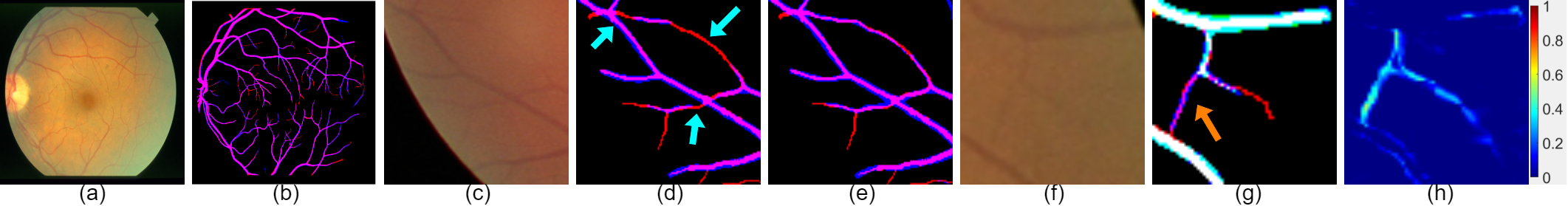}
    \centering
    \vspace*{-0.1in}
    \caption{In (a), (c), and (f), an example image and image regions are shown. In (b), (d), (e), and (g), Ground truth is shown in Red channel. In (b), our proposed framework output is shown in Blue channel. The magenta (red+blue) regions are true positives. Segmentation outputs of the model without weight maps (in (d)) and with weight maps (in (e)) are shown in Blue channel. Cyan arrows highlight the improved vessel connectivity obtained using vessel weight maps. In (g) segmentation outputs of the model without and with ILCs are shown in Green and Blue channels. Orange arrow highlights the improved vessel localization. In (h), the differences in network output probabilities (with\_ILCs $-$ without\_ILCs) are highlighted, showing that with ILCs, the network is more certain in segmenting low contrast vessels.}
    \label{fig:all_img}  
    \vspace*{-0.1in}
\end{figure*}

\subsection{Individual Objective Improvement}
\label{ssec:obj-imp}
We first present the details for improved tiny vessel segmentation using a vessel weight map based modified auxiliary loss calculation. Then we describe our proposed architecture for improved vessel localization.  

\begin{table*}[tb]
\caption{Segmentation results for the CHASE\_DB1, DRIVE, and STARE datasets.}
\small
\begin{center}
\scalebox{0.98}{
\begin{tabular}{ c | c  c  c  c | c  c  c  c | c  c  c  c}
\hline
\multicolumn{1}{c|}{} & \multicolumn{4}{c|}{CHASE\_DB1~\cite{chasedb1}} & \multicolumn{4}{c|}{DRIVE~\cite{drive}} & \multicolumn{4}{c}{STARE~\cite{stare}}\\
\hline
\rule{0pt}{8pt} Method & AUC & Acc & Spe & Sen & AUC & Acc & Spe & Sen & AUC & Acc & Spe & Sen \\ \hline
Fraz~\textit{et al.}~\cite{chasedb1} & 97.12 & 94.68 & 97.11 & 72.24 & 97.47 & 94.80 & 98.07 & 74.06 & 97.68 & 95.34 & 97.63 & 75.48 \\ 
Li~\textit{et al.}~\cite{li} & 97.16 & 95.81 & 97.93 & 75.07 & 97.38 & 95.27 & 98.16 & 75.69 & \textbf{98.79} & 96.28 & 98.44 & 77.26 \\
Orlando~\textit{et al.}~\cite{orlando} & 95.24 & - & 97.12 & 72.77 & 95.07 & - & 96.84 & 78.97 & - & - & 97.38 & 76.80 \\
Wu~\textit{et al.}~\cite{xia} & 98.25 & 96.37 & 98.47 & 75.38 & 98.07 & 95.67 & \textbf{98.19} & 78.44 \ & - & - & - & - \\
Mishra~\textit{et al.}~\cite{mishra} & 97.63 & 96.01 & 96.51 & 88.05 & 97.24 & 95.40 & 96.01 & 89.16 & 97.42 & 95.71 & 96.34 & 87.71 \\
Wu~\textit{et al.}~\cite{wu_new} & \textbf{98.94} & 96.88 & \textbf{98.80} & 80.03 & 98.30 & 95.82 & 98.13 & 79.96 & 98.75 & 96.72 & \textbf{98.63} & 79.63 \\
\hline \hline
Mou~\textit{et al.}~\cite{mou} & 98.12 & 96.37 & 97.73 & 82.68 & 97.96 & \textbf{95.94} & 97.88 & 81.26 & 98.58 & \textbf{96.85} & 97.69 & 83.91 \\
Ours & 98.63 & \textbf{97.18} & 97.78 & \textbf{89.07} & \textbf{98.33} & 95.84 & 96.50 & \textbf{90.14} & 98.59 & 96.66 & 97.23 & \textbf{89.11} \\
\hline
\end{tabular}}
\end{center}
\label{tab:chasedb1_drive_stare}
\vspace*{-0.1in}
\end{table*}

\noindent\textbf{Vessel Weight Map}: In order to preserve vessel structure, we propose vessel weight map to give some pixels more importance during training. Similar weight map inclusion was explored in \cite{unet} for learning border regions of cells which we modify and extend to vessel segmentation. Our weight map exerts a higher penalty based on the vessel topology (shown in Fig.~\ref{fig:overall}), forcing the network to yield connected vessels. For a label $o$ of a training image $i$, we perform distance transformation of $o$ to generate $d$. Using $d$, a pixel-wise weight map is computed as $wt\_map = \alpha \exp(-d^2/\beta^2)$, where $\alpha$ and $\beta$ are two hyperparameters. The weight map $wt\_map$ thus calculated is multiplied with the log softmax of the network output to generate the cross-entropy loss~\cite{unet}.

\noindent\textbf{Proposed Architecture}: We propose an encoder-decoder architecture as shown in Fig.~\ref{fig:mod-icc} for improved vessel segmentation. This architecture has a U-Net like backbone~\cite{unet} with a modified decoder (shown in dotted box in Fig.~\ref{fig:mod-icc}), where the side outputs (shown as thick colored arrows in Fig.~\ref{fig:mod-icc}) are merged to generate the main output, facilitating a multi-stage contextual information flow~\cite{cumednet}. To reduce false positive vessels and improve tiny vessel segmentation, we propose some improvements for enhanced vessel localization by introducing additional connections, called improved localization connections (ILCs). Each ILC acts as a support to the main input for a specific upsampling stage, resulting in enhanced localization. Intuitively, inclusion of feature maps of varying coarseness to generate the upsampling output is the key here. Similar inclusion of ILCs for each upsampling stage was used in~\cite{zhou_r} as redesigned skip paths. However, we modify it by proposing coarser \textit{stand-alone} upsampling paths without parameter sharing only for the \textit{target} upsampling stage.  Our approach of stand-alone upsamplers is based on the observation that robust latent feature learning for different image scales is not equivalent. Further, inclusion of ILCs in each upsampling stage introduces overfitting along with additional computational cost.

In order to introduce ILCs at the \textit{target} upsampling stage, consider a $CNN$ with $N$ stacked conv-layers in its encoder, arranged in $M$ stages. Each stage represents a set of conv-layers extracting features with a fixed feature map size (i.e., in between two downsampling operations). For a U-Net like backbone, there is a corresponding upsampling stage for each downsampling stage. Using the encoder's (downsampling stage) preeminent layer information determined in Section~\ref{ssec:unc_mod}, we locate its corresponding layer at the upsampling stage (or the target stage ($t$)) in the decoder. Let $y_m$ denote the output of the $m^{th}$ upsampler $U_m$. For U-Net, we obtain $y_m = U_m(x_m \oplus E_m)$, where $x_m$ is the input to $U_m$ and $E_m$ is the corresponding encoder stage's output. The ILCs are introduced for the target upsampling stage ($t$), making this stage's output be $y_t = U_t(x_t \oplus E_t \oplus U\!I_{t+1}(E_{t+1}) \oplus \cdots \oplus U\!I_{M}(E_M))$, where $E_i$ is the output of the encoder stage $i$, $i=t,\ldots,M$, $\oplus$ is the concatenation operation, and $U\!I_m$ is the transitional upsampling layer for ILCs used to upsample the deeper stage feature maps. For our experiments on retinal vessels, $t$ is determined to be the uppermost stage of the network (i.e., $t$ = 1), as shown in Fig.~\ref{fig:mod-icc}. Note that $U_1$ is a hypothetical layer performing trivial $1 \times 1$ upsampling, and $U\!I_M$ upsamples $E_M$ directly and does not contribute in the $y_{m-1}$ calculation.

\section{Experiments}
\label{sec:exp}

\textbf{Datasets:} We use three public retinal vessel segmentation datasets to evaluate our method. In the CHASE\_DB1~\cite{chasedb1} and STARE~\cite{stare} datasets, 28 and 20 retinal images are provided, respectively, without any specific training-test split. Following~\cite{mou}, we use a 4-fold cross-validation for experiments with CHASE\_DB1 and STARE datasets. The DRIVE~\cite{drive} dataset includes 20 training and 20 test retinal images. 

\noindent \textbf{Experimental Setup:} The experiments utilize the PyTorch framework with \textit{He} initialization~\cite{he_init}. To limit overfitting on a small training set, data augmentation is performed using random flipping and rotation. The training uses the Adam optimizer ($\beta_1 = 0.9, \beta_2 = 0.999, \epsilon = 1e-10$) with an initial learning rate 0.00005, which is halved in every $5k$ epochs for $20k$ epochs. Class balanced weights are used for cross-entropy calculation. The images are resized (CHASE\_DB1: 768 $\times$ 768, DRIVE: 512 $\times$ 512, STARE: 592 $\times$ 592), and the training uses 128$\times$128 size patches. For vessel map weight generation, $\alpha = 5$ and $\beta = 2$ are used in our experiments.

\noindent \textbf{Results:} Quantitative segmentation results obtained on the CHASE\_DB1, DRIVE, and STARE datasets are presented in Table~\ref{tab:chasedb1_drive_stare}. Improved sensitivity is observed for all the three datasets. Out of the recent works~\cite{mou,wu_new}, Mou~\textit{et al.}~\cite{mou} is considered as the state-of-the-art for comparison as it focuses tiny vessel segmentation. Improved AUC is observed for all three datasets over \cite{mou}. An example output obtained using our framework on the STARE dataset is shown in Fig.~\ref{fig:all_img}~(b). 

\noindent \textbf{Ablation Study:} In ablation study (shown in Table~\ref{tab:abl}), we explore the effect of our 3 major contributions. Experiments are conducted on the DRIVE dataset as an example case. \textit{\textbf{Uncertainty Modelling:}} Since we scale each loss using learned parameters ($\sigma_{main}$ and $\sigma_{aux}$), we compare our approach with static scaling. $\lambda = \{1,0.1,0.01,0.001\}$ are used to scale the auxiliary loss. In static scaling of the auxiliary loss, the network compromises sensitivity for AUC along with poor specificity. \textit{\textbf{Weight Map:}} In Table~\ref{tab:abl}, quantitative contribution of the vessel weight map is shown. An example highlighting better connectivity obtained with the vessel weight map is given in Fig.~\ref{fig:all_img}~(e). \textit{\textbf{ILCs:}} Contribution of our proposed encoder-decoder architecture with ILCs is shown in Table~\ref{tab:abl}. An example showing improved localization is given in Fig.~\ref{fig:all_img}~(g). We believe that introducing additional connections with shared weights for all decoder stages as in~\cite{zhou_r} incurs overfitting, causing reduction in accuracy.

\begin{table}[tb]
\vspace*{-0.1in}
\caption{Ablation study (\cmark: with and \xmark: without method).}
\small
\begin{center}
\scalebox{0.97}{
\begin{tabular}{ c  c  c  c  c  c  c}
\hline
\rule{0pt}{10pt} Uncert & Wt-map & ILCs & AUC & Acc & Sp & Se \\ \hline
$\lambda$ = 1 & \xmark & \xmark & 97.24 & 95.40 & 96.01 & 89.16\\ 
$\lambda$ = 1 & \xmark & \cmark & 97.86 & 95.52 & 96.12 & 89.49\\ 
$\lambda$ = 0.1 & \cmark & \cmark & 97.91 & 95.14 & 95.52 & 90.91\\ 
$\lambda$ = 0.01 & \cmark & \cmark & 98.06 & 95.41 & 95.92 & 90.20\\ 
$\lambda$ = 0.001 & \cmark & \cmark & 98.13 & 95.27 & 95.80 & 89.61\\ 
\cmark & \cmark & \cite{zhou_r} & 98.11 & 95.43 & 95.98 & 90.01 \\
\cmark & \xmark & \cmark & 98.08 & 95.65 & 96.26 & 89.63 \\
\cmark & \cmark & \cmark & 98.33 & 95.84 & 96.50 & 90.14 \\
\hline
\end{tabular}}
\end{center}
\label{tab:abl}
\vspace*{-0.1in}
\end{table}

\section{Conclusions}
\label{sec:conclusion}
In this paper, we proposed an objective-dependent uncertainty driven framework for retinal vessel segmentation by formulating vessel segmentation as accurate vessel segmentation and tiny vessel segmentation. Further, by introducing a vessel weight map based auxiliary loss to our proposed encoder-decoder architecture, we enforce vessel connectivity along with improved vessel localization. Comprehensive experiments demonstrate the efficacy of our method.

\section{Compliance with Ethical Standards}
\label{sec:ethics}
This research study was conducted retrospectively using human subject data made available in open access by \cite{chasedb1,drive,stare}. Ethical approval was not required as confirmed by the license attached with the open access data.

\section{Acknowledgement}
This work was supported in part by the National Science Foundation under Grants CNS-1629914, CCF-1640081, and CCF-1617735, and by the Nanoelectronics Research Corporation, a wholly-owned subsidiary of the Semiconductor Research Corporation, through Extremely Energy Efficient Collective Electronics, an SRC-NRI Nanoelectronics Research Initiative under Research Task ID 2698.004 and 2698.005.

\bibliographystyle{IEEEbib}
\balance
\bibliography{strings,refs}

\begin{thebibliography}{10}

\bibitem{chasedb1}
M.~{Fraz} et~al.,
\newblock ``An ensemble classification-based approach applied to retinal blood
  vessel segmentation,''
\newblock {\em IEEE Transactions on Biomedical Engineering}, 2012.

\bibitem{mou}
L.~{Mou} et~al.,
\newblock ``Dense dilated network with probability regularized walk for vessel
  detection,''
\newblock {\em IEEE Transactions on Medical Imaging}, 2020.

\bibitem{li}
Q.~{Li} et~al.,
\newblock ``A cross-modality learning for vessel segmentation in retinal
  images,''
\newblock {\em IEEE Transactions on Medical Imaging}, 2016.

\bibitem{drive}
J.~Staal et~al.,
\newblock ``Ridge based vessel segmentation in color images of the retina,''
\newblock {\em {IEEE Transactions on Medical Imaging}}, 2004.

\bibitem{orlando}
J.~{Orlando} et~al.,
\newblock ``A discriminatively trained fully connected conditional random field
  model for blood vessel segmentation in fundus images,''
\newblock {\em IEEE Transactions on Biomedical Engineering}, 2017.

\bibitem{avr}
P.~K. {Raj} et~al.,
\newblock ``Automatic classification of artery/vein from single wavelength
  fundus images,''
\newblock in {\em ISBI}, 2020.

\bibitem{xia}
Y.~Wu et~al.,
\newblock ``Multiscale network followed network for retinal vessel
  segmentation,''
\newblock in {\em MICCAI}, 2018.

\bibitem{yishuo}
Y.~Zhang et~al.,
\newblock ``Deep supervision with additional labels for retinal vessel
  segmentation task,''
\newblock in {\em MICCAI}, 2018.

\bibitem{wu_new}
Y.~{Wu} et~al.,
\newblock ``{NFN+}: {A} novel network followed network for retinal vessel
  segmentation,''
\newblock {\em Neural Networks}, 2020.

\bibitem{mishra}
S.~Mishra et~al.,
\newblock ``A data-aware deep supervised method for retinal vessel
  segmentation,''
\newblock in {\em ISBI}, 2020.

\bibitem{kendall}
A.~{Kendall} and Y.~{Gal},
\newblock ``What uncertainties do we need in bayesian deep learning for
  computer vision?,''
\newblock in {\em NIPS}, 2017.

\bibitem{cipolla}
R.~{Cipolla} et~al.,
\newblock ``Multi-task learning using uncertainty to weigh losses for scene
  geometry and semantics,''
\newblock in {\em CVPR}, 2018.

\bibitem{bischke}
B.~{Bischke} et~al.,
\newblock ``Multi-task learning for segmentation of building footprints with
  deep neural networks,''
\newblock in {\em ICIP}, 2019.

\bibitem{denker}
J.~{Denker} and Y.~{LeCun},
\newblock ``Transforming neural-net output levels to probability
  distributions,''
\newblock in {\em NIPS}, 1990.

\bibitem{stare}
A.~{Hoover} et~al.,
\newblock ``Locating blood vessels in retinal images by piecewise threshold
  probing of a matched filter response,''
\newblock {\em IEEE Transactions on Medical Imaging}, 2000.

\bibitem{unet}
O.~{Ronneberger} et~al.,
\newblock ``{U-Net}: Convolutional networks for biomedical image
  segmentation,''
\newblock {\em ArXiv e-prints}, May 2015.

\bibitem{cumednet}
H.~Chen et~al.,
\newblock ``Deep contextual networks for neuronal structure segmentation,''
\newblock in {\em AAAI}, 2016.

\bibitem{zhou_r}
Z.~{Zhou} et~al.,
\newblock ``{UNet++}: {A} nested {U-Net} architecture for medical image
  segmentation,''
\newblock {\em CoRR}, vol. abs/1807.10165, 2018.

\bibitem{he_init}
K.~He et~al.,
\newblock ``Delving deep into rectifiers: Surpassing human-level performance on
  {ImageNet} classification,''
\newblock {\em arXiv e-prints}, p. arXiv:1502.01852, 2015.

\end{thebibliography}

\end{document}


%
\maketitle
%


\begin{figure*}
    \vspace*{-0.1in}
    \includegraphics[width=17.2cm, height=14.2cm]{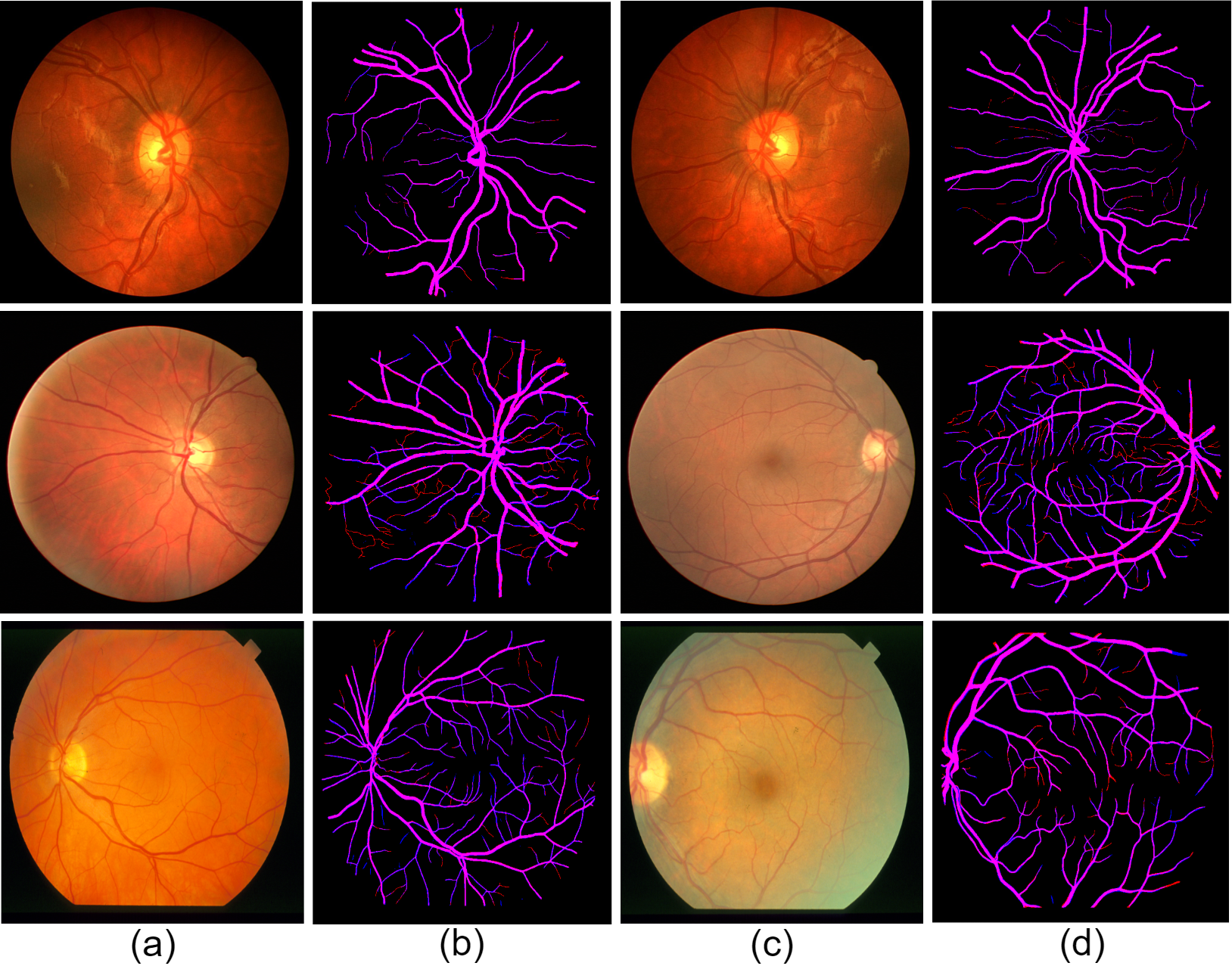}
    \centering
    \caption{Example fundus images are shown in (a) and (c). In (b) and (d), Ground truth is shown in Red channel, while segmentation outputs are shown in Blue channel. Magenta (red+blue) region, highlights the true positive pixels.}
    \label{fig:sample_results}  
\end{figure*}

\begin{figure*}
    \includegraphics[width=0.884\textwidth]{code_snippet.PNG}
    \centering
    \caption{ Code snippet for homoscedastic uncertainty capturing loss function.}
    \label{fig:sample_code}  
\end{figure*}
